\documentclass[10pt,twocolumn,letterpaper]{article}

\usepackage[preprint]{iccv}      

\definecolor{iccvblue}{rgb}{0.21,0.49,0.74}
\usepackage[pagebackref,breaklinks,colorlinks,allcolors=iccvblue]{hyperref}
\usepackage{xcolor}
\newcommand{\textgreen}[1]{\textcolor{green!50!black}{#1}}

\def\paperID{15668} 
\def\confName{ICCV}
\def\confYear{2025}

\title{\ours: Inference-Time Scaling for Text-to-Image Diffusion Transformers via In-Context Reflection}

\author{
Shufan Li$^{*1}$, Konstantinos Kallidromitis$^{*2}$, Akash Gokul$^{*3}$, Arsh Koneru${^1}$ \\ Yusuke Kato$^2$, Kazuki Kozuka $^2$, Aditya Grover$^1$  
\\ $^1$UCLA~ $^2$Panasonic AI Research~ $^3$Salesforce AI Research
\\
{ \tt\small *Equal Contribution }
\\
{ \tt\small Correspondence to jacklishufan@cs.ucla.edu}
}
\usepackage{tabularx}
\usepackage{algorithm}
\usepackage{algpseudocode}
\usepackage{tcolorbox}
\newcommand{\ours}{Reflect-DiT}
\newcolumntype{H}{>{\setbox0=\hbox\bgroup}c<{\egroup}@{}}

\begin{document}
\maketitle
\begin{abstract}
The predominant approach to advancing text-to-image generation has been training-time scaling, where larger models are trained on more data using greater computational resources. While effective, this approach is computationally expensive, leading to growing interest in inference-time scaling to improve performance. Currently, inference-time scaling for text-to-image diffusion models is largely limited to best-of-N sampling, where multiple images are generated per prompt and a selection model chooses the best output. Inspired by the recent success of reasoning models like DeepSeek-R1 in the language domain, we introduce an alternative to naive best-of-N sampling by equipping text-to-image Diffusion Transformers with in-context reflection capabilities. We propose \ours, a method that enables Diffusion Transformers to refine their generations using in-context examples of previously generated images alongside textual feedback describing necessary improvements. Instead of passively relying on random sampling and hoping for a better result in a future generation, \ours~explicitly tailors its generations to address specific aspects requiring enhancement. Experimental results demonstrate that \ours~improves performance on the GenEval benchmark (\textbf{+0.19}) using SANA-1.0-1.6B as a base model. Additionally, it achieves a new \textbf{state-of-the-art score of 0.81 on GenEval} while generating only 20 samples per prompt, surpassing the previous best score of 0.80, which was obtained using a significantly larger model (SANA-1.5-4.8B) with 2048 samples under the best-of-N approach.\footnote{Code will be available at \href{https://github.com/jacklishufan/Reflect-DiT}{https://github.com/jacklishufan/Reflect-DiT}}
\end{abstract}
    
\begin{figure}[t]
    \centering
    \includegraphics[width=1\linewidth]{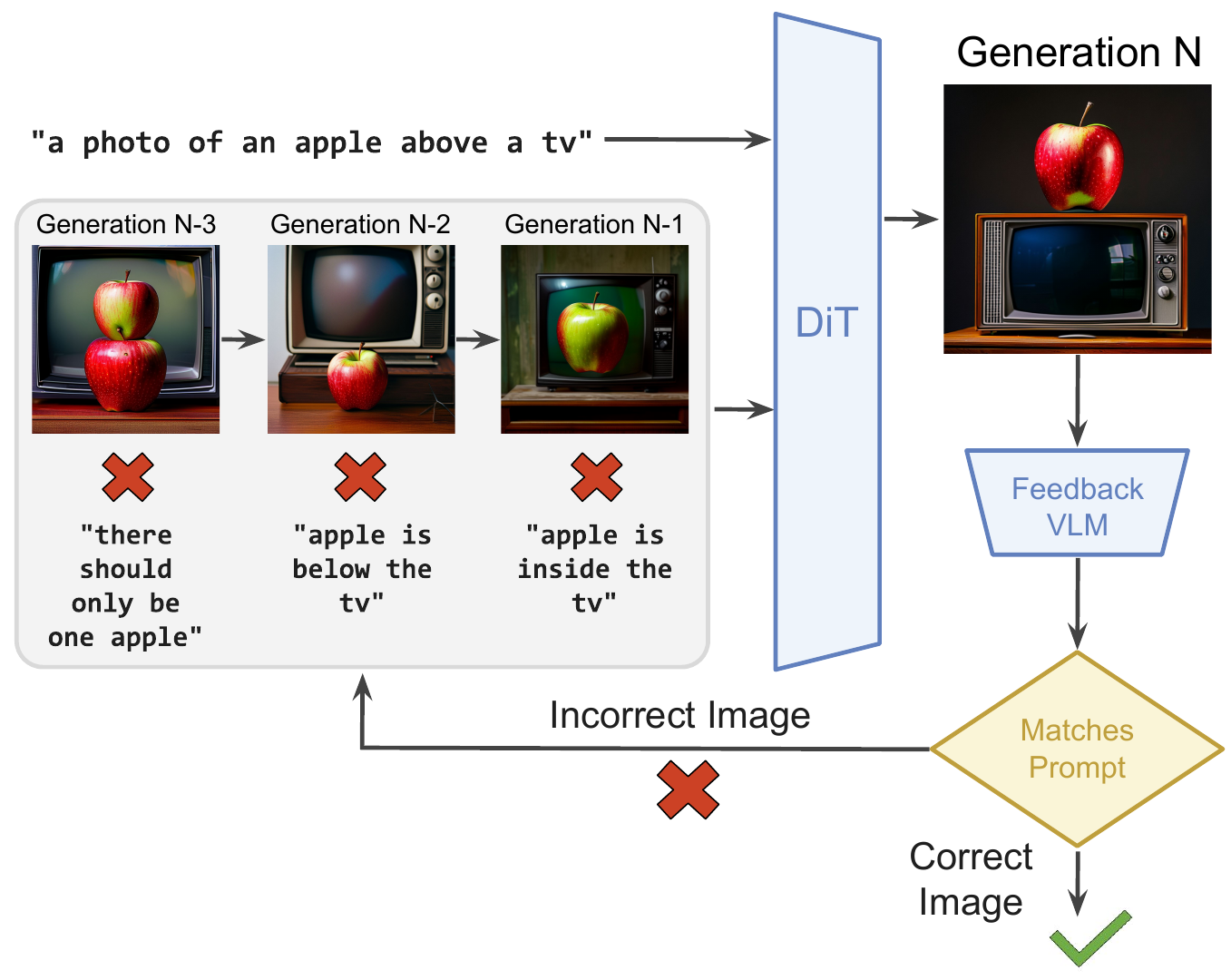}
    \caption{\textbf{\ours~iteratively refines image generation by using a vision-language model (VLM) to critique generations and a Diffusion Transformer (DiT) to self-improve using past generations and feedback.} Specifically, at each generation step N, feedback from previous iterations (N-3, N-2, N-1, \dots) are incorporated to progressively improve future generations. Unlike traditional best-of-N sampling, \ours~actively corrects errors in object count, position, and attributes, enabling more precise generations with fewer samples.}
    \label{fig:figure-teaser}
\end{figure}

\section{Introduction}
\label{sec:intro}
%
%

Text-to-image diffusion models have made significant progress by training larger architectures with more data \cite{rombach2022high, podell2023sdxl, ramesh2021zero, xie2025sana1}. However, scaling the training of such models, \ie using more data and/or larger models, is computationally expensive, as exponential increases in compute are required to achieve near-linear performance gains \cite{liang2024scaling}. Recently, works studying Large Language Models (LLMs), which exhibit similar scaling laws, have explored inference-time scaling as another means of improving model performance \cite{brown2024large,wu2024inference,snell2024scaling,zhang2024generative}. These approaches introduce additional compute resources during inference, either by generating many samples and employing a selection mechanism to find the best output \cite{lightman2023let} or by prompting the LLM to generate longer reasoning traces, often involving self-verification and self-correction, before reaching an answer \cite{weng2022large,chen2025sets}. These methods have shown promising results, offering substantial performance gains with a relatively moderate increase in compute compared to training-time scaling. 

More recently, several works have attempted to apply the concept of inference-time scaling to text-to-image diffusion models \cite{xie2025sana1,ma2025inference}. They primarily focus on two key aspects: scaling denoising steps per sample and scaling the number of samples. For the latter, a commonly used framework is to generate N random samples per prompt and select the best result using a reward or judge model (best-of-N). In particular, SANA-1.5 \cite{xie2025sana2} achieved state-of-the-art performance on the GenEval benchmark \cite{ghosh2023geneval} by generating a very large number of samples per prompt (N=2048) using a 4.8B Diffusion Transformer (DiT). Its performance surpassed previous results achieved by significantly larger models, highlighting the benefits of inference-time scaling.

Despite their encouraging success, these methods still have considerable room for improvement in terms of efficiency. For example, generating 2048 samples per prompt is impractical for real-world applications. In this work, we propose \ours, an effective framework to improve the inference-time scaling of Diffusion Transformers by equipping DiTs with the ability to refine future generations through reflecting upon its past generated images and natural language feedback. \ours~draws inspiration from the recent success of reasoning models, such as DeepSeek-R1 \cite{guo2025deepseek}, which exhibit self-verification and reflection capabilities. Rather than relying on random sampling to produce a better output in the next generation, these models utilize their long context windows to reason about a given problem. This process allows the model to search, reflect on, and refine potential solutions before providing a final output. Unlike autoregressive LLMs, text-to-image diffusion models currently lack the ability to reason about past generations and feedback, as they condition solely on the input prompt. We argue that this limitation—specifically, the inability to reference and learn from past generations—prevents them from achieving inference-time scaling benefits beyond naive best-of-N sampling strategies. To address this shortcoming, \ours~incorporates in-context reflection, expanding the conditioning signals to include past image generations and natural language feedback (\cref{fig:figure-teaser}). \ours~can evaluate its past generations, identify misalignments with the input prompt (\eg object count, spatial positioning, \etc) and refine subsequent generations to correct these issues.

Concretely, \ours~consists of (1) a Vision-Language Model (VLM) that serves as a judge that evaluates generated images with respect to input prompts and provides natural language feedback, and (2) a Diffusion Transformer that refines its generations based on previous generations and corresponding feedback. Previously generated images and text feedback are first encoded with vision and text encoders into modality-specific embedding spaces, then processed by a lightweight Context Transformer to obtain a set of conditional embeddings that are passed to the cross-attention layers of the DiT. To ensure scalability, \ours~maintains a fixed context length, limiting the number of past generations it considers. When the total number of past generations exceeds the context limit, we employ a selection mechanism to stochastically sample a subset of past generations as the context. The \ours~framework is illustrated in Figure \ref{fig:figure-teaser}.

We conduct extensive experiments to test the effectiveness of \ours. Compared to the naive best-of-N approach, \ours~improves performance by \textbf{+0.19 on the GenEval benchmark} using SANA-1.0-1.6B as the base model. \ours~also establishes a new absolute \textbf{state-of-the-art score of 0.81 on GenEval} while generating only 20 samples per prompt during inference, surpassing the previous best score of 0.80 that was obtained using a significantly larger model (SANA-1.5-4.8B) with 2048 samples under the naive best-of-N approach. These results highlight \ours~as a more effective and efficient alternative to best-of-N sampling for inference-time scaling of DiTs.

\begin{figure*}[t]
    \centering
    \includegraphics[width=1.0\linewidth]{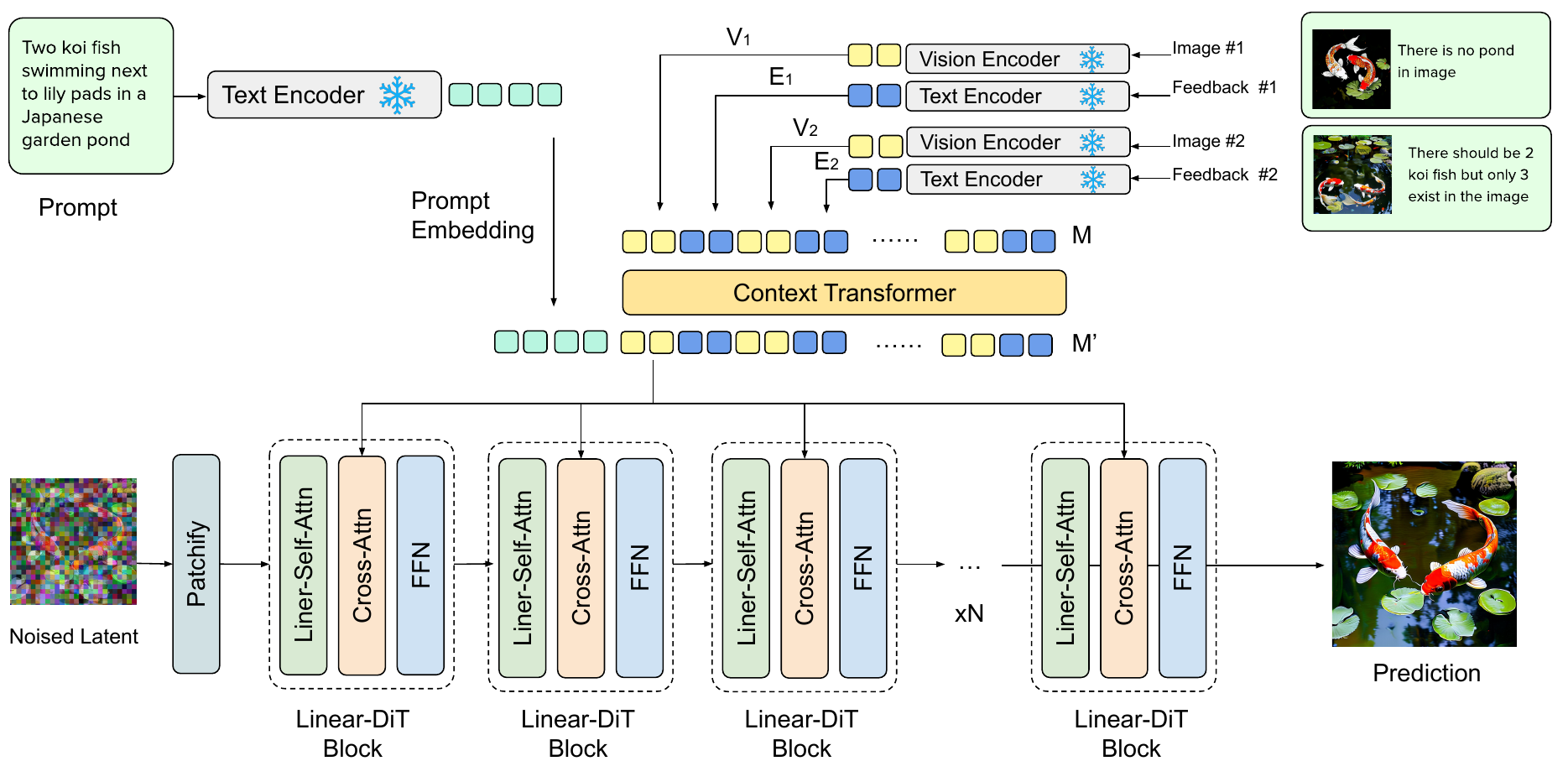}
    \caption{\textbf{Architecture of \ours}. Given a prompt, past images and feedback, we first encode the images into a set of vision embeddings $[V_1,V_2,\dots]$ using a vision encoder, and encode text feedback to a set of text embeddings $[E_1,E_2...]$. We then concatenate these embeddings into a single sequence $M$, and pass it through the Context Transformer to obtain $M'$. The extra context $M'$ is concatenated directly after the standard prompt embeddings and passed into the cross-attention layers of the Diffusion Transformer (DiT).
    }
    \label{fig:explainer}
\end{figure*}

\section{Related Works}

\subsection{Inference-time scaling of LLMs}
Traditionally, stronger performance on language tasks was achieved by training larger models using more data \cite{brown2020gpt3,openai2023gpt4}. Recent work has begun to explore using additional compute at test time to improve performance. \citet{brown2024large} demonstrated the benefits of combining repeated sampling and a selection mechanism such as automatic verifiers or reward models. \citet{snell2024scaling} discovered that adaptive search and iterative self-refinement are more effective than naive Best-of-N sampling given a fixed compute budget. \citet{wu2024inference} showed that with proper inference strategies, scaling test-time compute can be more efficient than scaling model parameters. Most recently, several works \cite{chen2025sets,ma2025s} showed promising results in improving LLMs' performance by spending additional compute for self-verification and self-correction at test time. Building on these insights, \ours~explores inference-time scaling beyond naive Best-of-N sampling by enabling text-to-image models to iteratively refine their generations by reflecting on past generations and feedback.

\subsection{Inference-time scaling of Text-to-Image Diffusion Models}

Inspired by the success of inference-time scaling of LLMs, recent works have explored similar strategies for text-to-image diffusion models. \citet{ma2025inference} explored several search strategies and concluded that random search combined with Best-of-N selection is currently the best strategy for improving performance. SANA-1.5 \cite{xie2025sana2} applied Best-of-N sampling to a frontier text-to-image model and established a new state of the art on the GenEval benchmark through Best-of-N sampling (N=2048). Concurrent to this work, \citet{singhal2025general} proposed a more structured search method that extends beyond random sampling.
 
\subsection{Self-Correction for Text-to-Image Generation}
Before the recent advancements in inference-time scaling for text-to-image models, earlier works explored using LLM agents to enable self-verification and self-improvement (SLD\cite{wu2024self} and GenArtist\cite{wang2024genartist}). These methods use LLMs or VLMs as controllers to generate a series of image operations or function calls to specialized models \cite{liu2024grounding, kirillov2023segment, yu2023inpaint, Brooks_2023_CVPR}. Unlike \ours, these works do not enable text-to-image diffusion models to learn from natural feedback; instead, they rely on a predefined and limited set of operations. (\eg object manipulation in latent space \cite{wu2024inference}). Furthermore, the performance of SLD and GenArtist is highly dependent on the successful execution of each submodule/operation, many of which involve heuristics or are not up-to-date. Because of such limitations, these approaches are not ideal for inference time scaling of frontier text-to-image models. We consider these works to be tangentially related to \ours~rather than directly comparable. Further discussion can be found in the Appendix \ref{sec:appendix-agent}.

\subsection{Controllable Text-to-Image Generation}

Text-to-image generation is inherently constrained by language as it relies solely on the input prompt. Thus, limiting its ability to represent concepts that may be difficult to capture in words. In the case of subject-driven generation, \ie generating an image of a specific subject such as one's pet, conditioning solely on text has shown to be ineffective. To address this, subject-driven generation methods introduce learnable embeddings \cite{gal2022image, ruiz2023dreambooth, kumari2023multi, voynov2023p+}, or provide visual conditioning signals \cite{li2023blip, wei2023elite, ma2024subject, purushwalkam2024bootpig}. Furthermore, previous works \cite{zhang2023adding, avrahami2023spatext, xie2023boxdiff, mou2024t2i, rombach2022text, sheynin2022knn, zhao2023uni, li2023gligen,chenre, huang2023composer} have introduced mechanisms to expand the conditioning signal of text-to-image diffusion models to include a variety of signals to enable control beyond text. Unlike \ours, these works focus on zero-shot controllable image generation and do not use conditioning signals to allow the model to learn from iterative feedback. Prior works have also explored using natural language instructions to improve image editing \cite{Brooks_2023_CVPR,geng2024instructdiffusion,zhang2024hive,jin2024reasonpix2pix,Feng_2024_CVPR}. We provide further comparisons with these works in Section \ref{sec:comparision_with_instructpix2pix}.

\section{Method}

\subsection{Overall Framework}
Similar to its counterparts in the language domain, \ours~iteratively refines its generation by performing a verification-reflection loop. \ours~consists of a vision-language model (VLM) feedback judge $F_{j}$ that generates text feedback $T$ for an input image $X$, and a Diffusion Transformer (DiT) text-to-image generator $F_{g}$ that maps a text prompt $P$ and a set of past generations and feedback $C = \{(X_i, T_i) \mid i = 1, 2, \dots\}$ to a new output image $X_j$. Given an input prompt $P$, we first generate an image $X_0=F_g(P)$ without any additional context, and obtain the initial feedback $T_0=F_j(P,X_0)$. At each subsequent iteration $i$, we obtain a new generation $X_i$ and its corresponding feedback $T_i$. The reflection context is then updated, as $C_i = \{(X_j, T_j) \mid j = 1, 2, \dots, i\}$, to include all past generations and feedback. If the size of $C_i$ is larger than a pre-defined max context length $K$, we randomly sample $K$ past generations and their corresponding feedback as the input context $C_k$. Otherwise, we use all pairs in $C_i$ to form input context $C_k$, The generator then produces an updated image $X_{i+1} = F_g(P, C_k)$, incorporating past feedback. This loop continues until the VLM feedback judge produces null feedback (indicating no further improvements) or a maximum number of iterations $N$ is reached. The full procedure is formally described in \cref{alg:inference}.

\subsection{VLM Feedback}
The goal of the VLM judge, $F_{j}$, is to provide natural language feedback for generated images. We use Qwen2.5-VL 3B\cite{bai2025qwen2} as the judge model and finetune it following the setup of SANA 1.5 \cite{xie2025sana2}. Specifically, the training data is curated by generating a large number of images from synthetic prompts and using an object detector to judge whether the desired objects are present in the image and whether their counts and attributes agree with the prompt. Feedback data used in VLM training are generated using structured templates. For example, if an object X is missing, we use the template ``There is no \{X\} in the image." If the count of objects is incorrect, we use the template ``There should be \{N\} \{X\} in the image, but only \{K\} exist." The feedback is intentionally concise to improve VLM inference efficiency and minimize memory overhead for the DiT. We provide additional details of VLM training in Appendix \ref{sec:appendix-vlm-training}. 

\begin{algorithm}[t]
    \caption{Iterative Image Refinement with Verification-Reflection Loop}
    \label{alg:inference}
    \begin{algorithmic}[1]
        \Require Text prompt $P$, Feedback Judge VLM $F_j$, DiT Image Generator $F_g$, Max Context Length $K$, Max Iterations $N$
        \State Initialize $X_0 \gets F_g(P)$  \Comment{Generate initial image}
        \State $T_0 \gets F_j(P, X_0)$  \Comment{Obtain initial feedback}
        \State $C_0 \gets \{(X_0, T_0)\}$  \Comment{Initialize reflection context}
        \For{$i = 1$ to $N$}
            \If{$T_{i-1} = \emptyset$}  \Comment{Stop if no more improvements}
                \State \textbf{break}
            \EndIf
            \State Construct $C_i = \{(X_j, T_j) | j = 1, 2, ..., i\}$
            \If{$|C_i| > K$}  
                \State Sample $K$ elements from $C_i$ to obtain $C_k$
            \Else
                \State $C_k = C_i$
            \EndIf
            \State Generate new image $X_i \gets F_g(P, C_k)$
            \State Obtain feedback $T_i \gets F_j(P, X_i)$
            \State Update reflection context $C_i \gets C_i \cup \{(X_i, T_i)\}$
        \EndFor
        \State \Return Image Trajectory $\{X_0, X_1, ..., X_n\}$
    \end{algorithmic}
\end{algorithm}

\begin{table*}[t]
    \centering
    \begin{tabular}{c | c H|c|cccccc}
        \textbf{Generator} &\textbf{Params} & \textbf{Judge} & \textbf{Overall} & \textbf{Single} & \textbf{Two} & \textbf{Counting} & \textbf{Color} & 
        \textbf{Position} & \textbf{Attribution} \\
        \hline
        SDXL\cite{podell2023sdxl} & 2.6B & - & 0.55 & 0.98 & 0.74 & 0.39 & 0.85 & 0.15 & 0.23 \\
        DALLE 3\cite{betker2023improving} & - & - & 0.67 & 0.96 & 0.87 & 0.47 & 0.83 & 0.43 & 0.45 \\
        SD3\cite{esser2024scaling} & 8B& - & 0.74 & 0.99 & 0.94 & 0.72 & 0.89 & 0.33 & 0.60 \\
        Flux-Dev\cite{flux2024} & 12B& - & 0.68 & 0.99 & 0.85 & 0.74 & 0.79 & 0.21 & 0.48 \\
        Playground v3\cite{liu2024playground} &- & - & 0.76 & 0.99 & 0.95 & 0.72 & 0.82 & 0.50 & 0.54 \\
        \hline
        SANA-1.5-4.8B\cite{xie2025sana2}  $\ddag$ & 4.8B  & - & 0.76 & 0.99 & 0.95 & 0.72 & 0.82 & 0.50 & 0.54 \\
        + Best-of-2048 $\ddag$ & 4.8B  & - & 0.80 & \textbf{0.99} & 0.88 & 0.77 & \textbf{0.90} & 0.47 & \textbf{0.74}  \\
        \hline
        SANA-1.0-1.6B $\dag$ \cite{xie2025sana1} & 1.6B & - & 0.62 & 0.98 &  0.83 & 0.58 & 0.86 &  0.19  &  0.37 \\
        + Best-of-20 &   1.6B  & - & 0.75 & 0.99 & 0.87 & 0.73 & 0.88 & 0.54 & 0.55 \\
       \textbf{+ \ours(N=20)} &   1.6B + 0.1B  & - & \textbf{0.81} & 0.98 & \textbf{0.96} & \textbf{0.80} & 0.88 & \textbf{0.66} & 0.60 \\
       \textbf{ ($\Delta$ vs Baseline)} & - & - & \textgreen{\textbf{+0.19}} & \textcolor{gray}{\textbf{+0.00}} & \textgreen{\textbf{+0.13}} & \textgreen{\textbf{+0.22}} & \textgreen{\textbf{+0.02}} & \textgreen{\textbf{+0.47}} & \textgreen{\textbf{+0.23}} \\
       \hline
        
    \end{tabular}
    \caption{\textbf{Results on the GenEval benchmark \cite{ghosh2023geneval}}. \ours~achieves the highest overall score (0.81) with only 20 samples per prompt, outperforming all other models despite having significantly fewer parameters. Compared to the base SANA-1.0-1.6B, \ours~demonstrates consistent improvements across all evaluation categories, with a notable overall gain of +0.19. While SANA-1.5-4.8B achieves competitive performance, it requires substantially more computational resources and is not open-sourced at the time of writing. $\dag$ Evaluated using the released checkpoint of SANA-1.0. $\ddag$ SANA-1.5 is not open-sourced; results are reported from the original paper.}
    \label{tab:results-geneval}
\end{table*}

\subsection{Diffusion Transformer}
We implement \ours~using SANA-1.0-1.6B \cite{xie2025sana1}, which offers competitive performance while being 106$\times$ faster than open-source alternatives\cite{podell2023sdxl, flux2024}. This improvement in efficiency is due to its smaller size and adoption of a linear attention mechanism, making it an ideal choice for inference-time scaling which requires generating a large number of samples per prompt. SANA consists of consecutive Linear-DiT blocks, each containing a self-attention layer, a cross-attention layer, and a feed forward network (FFN). We incorporate past generated images and text feedback as additional context for the cross-attention layer. 

Concretely, given prompt $P$ and past context $C=\{(X_1,T_1),(X_2,T_2),\dots\}$, we first encode the images $[X_1, X_2, \dots]$ into a sequence of vision embeddings $[V_1,V_2,\dots]$ using a vision encoder. We use SigLIP-Large \cite{zhai2023sigmoid} as our vision encoder, which encodes each image to a feature map of size $24 \times 24$. We downsample this feature map to $8 \times 8$ before flattening it into a 1D sequence, reducing the sequence length of image embeddings from 576 to 64. This compression minimizes the additional memory needed, making it computationally feasible to fit multiple images in the context during training and inference. To encode text feedback, we use Gemma-2-2B \cite{team2024gemma} and convert the text feedback $[T_1,T_2...]$ to a set of 1D embeddings $[E_1,E_2...]$. Finally, we concatenate these image and text embeddings into a single sequence $M=\text{Concat}([V_1,E_1,V_2,E_2,...])$. Since each image and text feedback are encoded separately, these features are not aligned with the features of the base DiT and there is no mechanism to associate each image with its corresponding feedback. Hence, we first process $M$ through a small two-layer Transformer \cite{vaswani2017attention}, which we refer to as the Context Transformer, to obtain intermediate output $M'$. The extra context $M'$ is concatenated directly after the standard prompt embeddings of SANA-1.0 and passed into the cross-attention layers of the DiT. This is illustrated in Figure \ref{fig:explainer}.

The same synthetic dataset is used to train both the VLM judge and the Diffusion Transformer. Specifically, each training data sample used for training the DiT consists of a ``good" image $X_w$ as the positive sample, and a set of ``bad" images $[X_1,X_2,...]$ and corresponding feedback $[T_1,T_2,...]$ as reflection context. The VLM training data consists of multiple image-feedback pairs per prompt. We select images that pass the object detector test (and hence use the feedback template ``This image is correct") as positive samples and the remaining images as in-context feedback samples. Prompts that lead to only ``good" images or ``bad" images are excluded from the training data. Following SANA, we finetune the DiT using the flow-matching objective as follows:

    \begin{align}
      \mathbb{E}_{t\in \text{Unif}[0,1]} w(t)\lVert\epsilon - x_w- F_g(x_w^t,t,C) \rVert^2
    \end{align}
where $\epsilon$ is an i.i.d Gaussian noise, $x_w$ is the image latent corresponding to the ``good" image $X_w$ and $x_w^t=(1-t)x_w+t\epsilon$ is a noised image latent, $C$ is the set of in-context image-feedback pairs, and $w(t)$ is a weighting function. We use a logit-normal weighting scheme following SANA.



\begin{figure*}[t]
    \centering
    \includegraphics[width=1.0\linewidth]{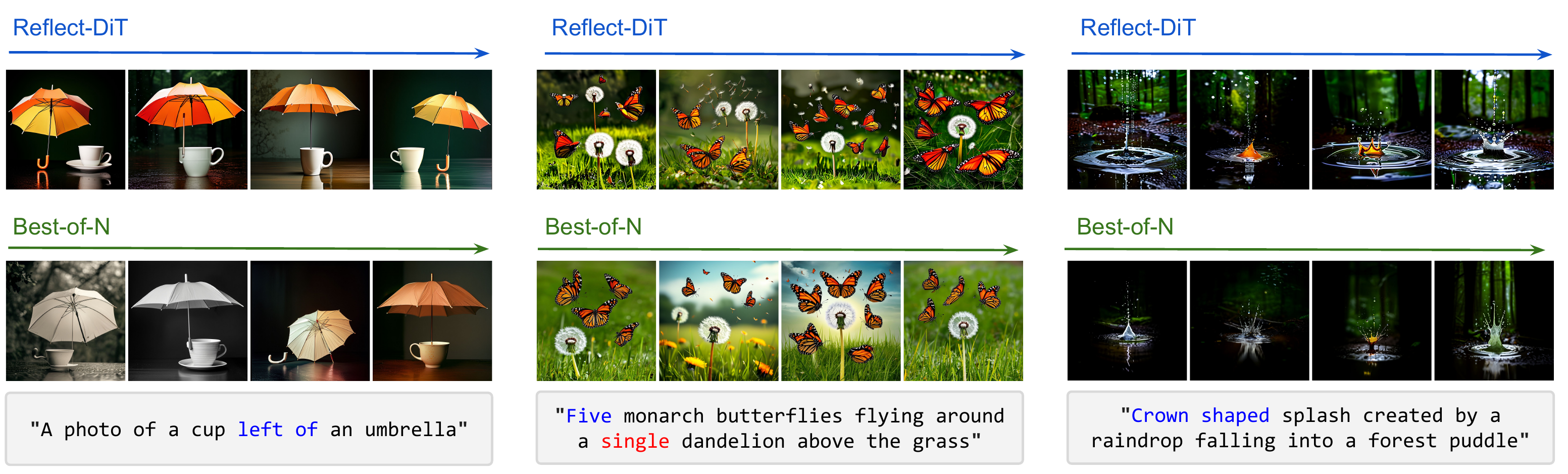}
    \caption{\textbf{Side-by-side qualitative comparison of \ours~and best-of-N sampling.} \ours~leverages feedback to iteratively refine image generations, resulting in more accurate and visually coherent outputs. In the first example, \ours~progressively adjusts object positions to better satisfy the prompt ``a cup \emph{left of} an umbrella," achieving significantly better image-text alignment than best-of-N sampling. The second example demonstrates how \ours~corrects multiple counting constraints (``\emph{five} monarch butterflies" and ``\emph{a single} dandelion") over successive iterations, gradually converging to the correct solution. Lastly, in the rightmost example, \ours~uses in-context feedback to refine object shapes, producing a more precise and intentional design compared to best-of-N.}
    \label{fig:sid-by-side}
\end{figure*}

\section{Experiments}
\subsection{Setup}
\subsubsection{Dataset}
We generate prompts using the GenEval templates \cite{ghosh2023geneval} and filter out those present in the test set, resulting in 6,000 prompts. We generate 20 images per prompt and obtain synthetic feedback using object detectors \cite{ghosh2023geneval}. This results in a dataset of 78.5k image-feedback pairs. Our pipeline follows the setup of SANA-1.5 but produces a smaller dataset (2M images in SANA-1.5) due to computational constraints.

\subsubsection{Training}
We train the VLM judge for 1 epoch with a learning rate of 1e-5. We train the DiT and Context Transformer for 5,000 steps with a learning rate of 1e-5 and batch size of 48. We freeze the image and text encoders, and finetune the DiT and Context Transformer end-to-end. Training is conducted on Nvidia A6000 GPUs and takes approximately one day.

\subsection{Sampling}

For all experiments, we use the DPM-Solver++ sampler proposed by SANA. We use 20 sampling steps per image and set the maximum number of images per prompt to 20 (N=20). For \ours, we set the maximum number of in-context feedback to 3 (K=3) unless stated otherwise. We provide further details on inference speed in Appendix \ref{sec:appendix-speed}.

\subsubsection{Baselines}
We compare against several training-free baselines and finetuning-based methods. For training-free baselines, we compare against best-of-N sampling employed by SANA-1.5. For finetuning baselines, we consider supervised finetuning (SFT) and Diffusion-DPO\cite{wallace2024diffusion}. For both methods, we use SANA-1.0-1.6B as the base model. Since these finetuning methods do not enable additional inference-time scaling capabilities, we combine them with best-of-N sampling to equalize test-time compute.

\subsection{Main results}

We report results on the GenEval benchmark in Table \ref{tab:results-geneval}. \ours~achieves the highest overall score (0.81) using at most 20 samples per prompt, outperforming all other models, including those with substantially more (3$\times$) parameters. Compared to the SANA-1.0-1.6B baseline, \ours~demonstrates consistent improvements across all evaluation categories, with an overall gain of +0.19. Notably, we establish a new state-of-the-art (0.81), surpassing the previous best (0.80), which was achieved using best-of-N sampling with N=2048 using SANA-1.5-4.8B. Improvements are especially pronounced for prompts requiring complex reasoning over multiple objects (\eg counting, spatial positioning, and attribute binding). As prompt complexity increases, the baseline model struggles to generate high-quality samples, since the probability of satisfying multiple constraints is low.

\begin{figure}[t]
    \centering
    \includegraphics[width=1.0\linewidth]{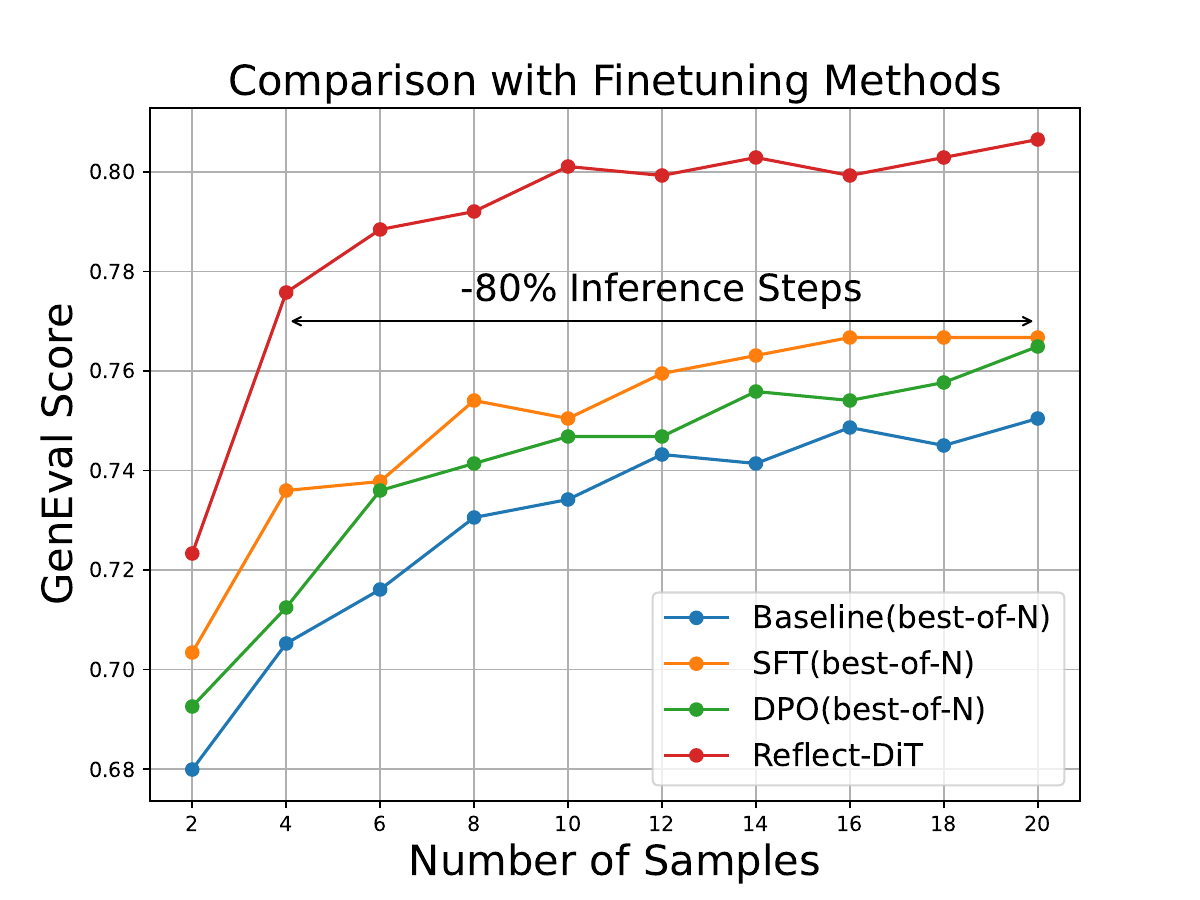}
    \caption{\textbf{Comparison of \ours~with other finetuning methods}. We find that \ours~ is able to consistently outperform finetuning methods, like supervised finetuning (SFT) and Diffusion-DPO (DPO). Using only 4 samples, \ours~can outperform related finetuning baselines using best-of-20 sampling.}
    \label{fig:finetune}
\end{figure}

\subsection{Qualitative Examples}
We present samples from the generation trajectory of \ours~in Figure \ref{fig:sid-by-side}. Compared with best-of-N sampling with random search, \ours~improves image quality more efficiently by tailoring future generations according to prior feedback. For example, in the rightmost trajectory of Figure \ref{fig:sid-by-side} the initial generations fail to render a ``crown shaped splash", but as the \ours~trajectory progresses, the rendered splash resembles the shape of a crown (top row). In contrast, best-of-N sampling (bottom row) fails to find a generation with a crown shaped splash. 

Figure \ref{fig:figure-feedback} showcases some examples of the self-correction process of \ours. The vision-language model correctly identifies misalignments between prompts and generated images, enabling \ours~to improve subsequent generations based on this feedback. For example, the topmost row of Figure \ref{fig:figure-feedback} shows generations from a prompt involving multiple entities and relative positioning constraints. Here, the VLM judge accurately identifies issues in generated images, such as inconsistent positioning and missing subjects, and \ours~is able to improve future generations based on this feedback. 


\begin{table}[t]
    \centering
    \begin{tabular}{lccccc}
        \toprule
         & \multicolumn{5}{c}{Number of Samples }\\
         & 4 & 8 & 12 & 16 & 20 \\
        \midrule
        Baseline & 0.705 & 0.731 & 0.743 & 0.749 & 0.751 \\
        +SFT & 0.736 & 0.754 & 0.760 & 0.767 & 0.767 \\
+DPO & 0.713 & 0.741 & 0.747 & 0.754 & 0.765 \\
        +Refl. DiT & \textbf{0.776} & \textbf{0.792} & \textbf{0.799} & \textbf{0.799} & \textbf{0.807} \\
        \bottomrule
    \end{tabular}
    \caption{\textbf{GenEval performance using different finetuning methods.} Results show that \ours~consistently outperforms supervised finetuning (SFT), and Diffusion-DPO (DPO) across a varying number of samples at inference. We use best-of-N sampling for the base model, SFT, and DPO baselines and in-context reflection for \ours.}
    \label{tab:finetuning-result}
\end{table}

\subsection{Comparison with Finetuning}

We additionally compare with several finetuning methods, including supervised-finetuning (SFT) and Diffusion-DPO \cite{wallace2024diffusion}. We report results on GenEval with varying number of samples in Table \ref{tab:finetuning-result} and Figure \ref{fig:finetune}. Since SFT and DPO do not enable additional inference-time scaling, we combine them with best-of-N sampling. \ours~outperforms both baselines by a considerable margin, especially at large number of samples. Using just 4 samples, \ours~outperforms both the SFT and DPO baselines with 20 samples, saving 80\% of the compute budget. Among the baselines, SFT and DPO have nearly identical performance with N=20 samples, but SFT outperforms DPO on fewer samples, presumably because of the KL-divergence penalty.

\subsection{Human Evaluation}
\begin{figure}[t]
    \centering
    \includegraphics[width=1.0\linewidth]{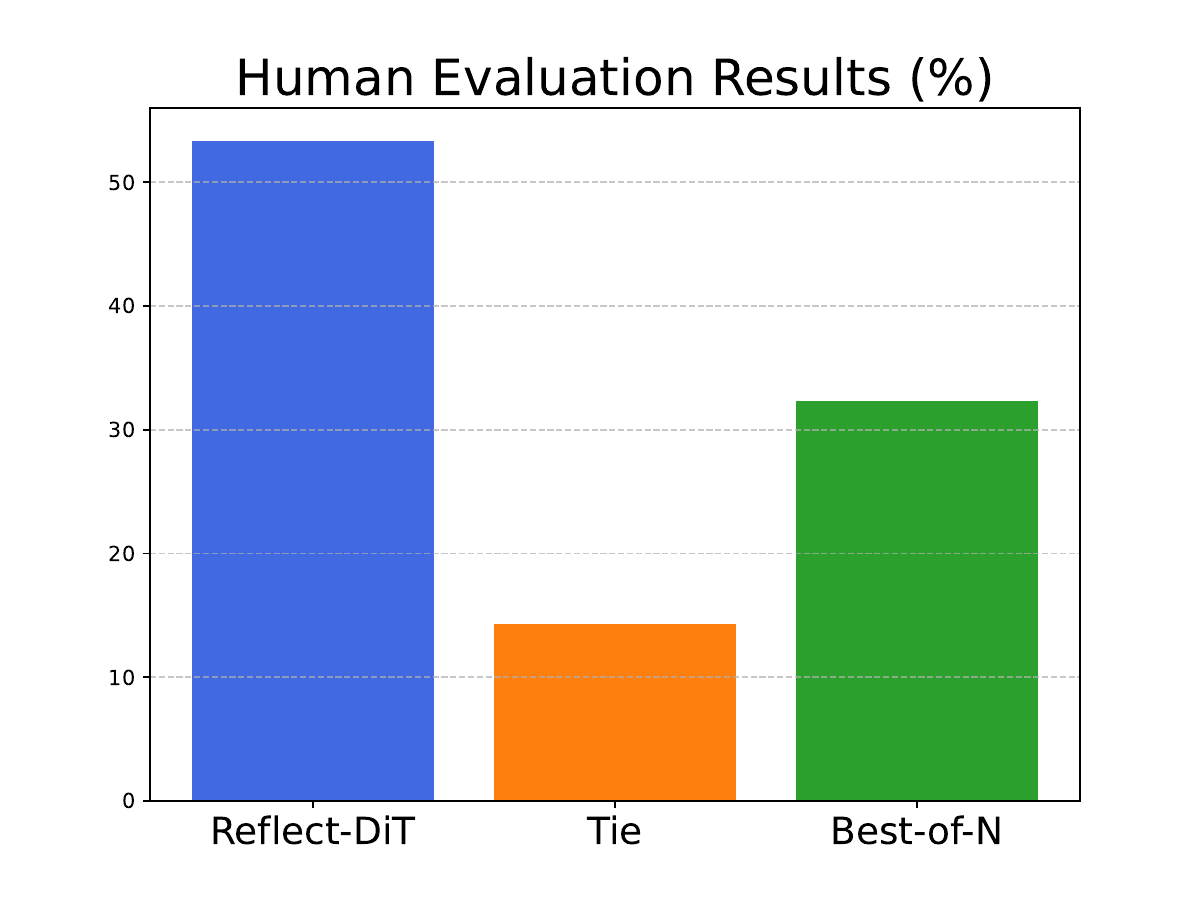}
    \caption{\textbf{Human evaluation win-rate (\%) on PartiPrompts dataset}. We perform a user study to evaluate the effectiveness of \ours~in broadly improving text-to-image generation. Results show that human evaluators consistently prefer generations from \ours~over best-of-N sampling.}
    \label{fig:human-eval}
\end{figure}

To evaluate \ours's effectiveness in real-world use cases beyond GenEval-style prompts, we conduct additional evaluations using 100 randomly sampled prompts from PartiPrompts\cite{yu2022scaling}. We conduct human evaluations and ask evaluators to compare \ours~against best-of-N sampling, with a maximum of 20 samples per prompt for each method. We attempted to evaluate generations using frontier VLMs such as GPT-4o, but encountered similar problems as \cite{xie2025sana2}, where the API gives inconsistent outputs and exhibits a strong bias towards the first presented image. Evaluators are given a pair of images and a prompt and asked to determine which generated image is better, without any prior knowledge of how they were generated. Three responses are collected per image pair. Our results demonstrate that \ours~significantly outperforms the best-of-N baseline, with evaluators selecting \ours~53.3\% of the time compared to 32.3\% for best-of-N (\cref{fig:human-eval}). We provide additional details in Appendix \ref{sec:appendix-human-eval}.

\begin{figure*}[t]
    \centering
    \includegraphics[width=1.0\linewidth]{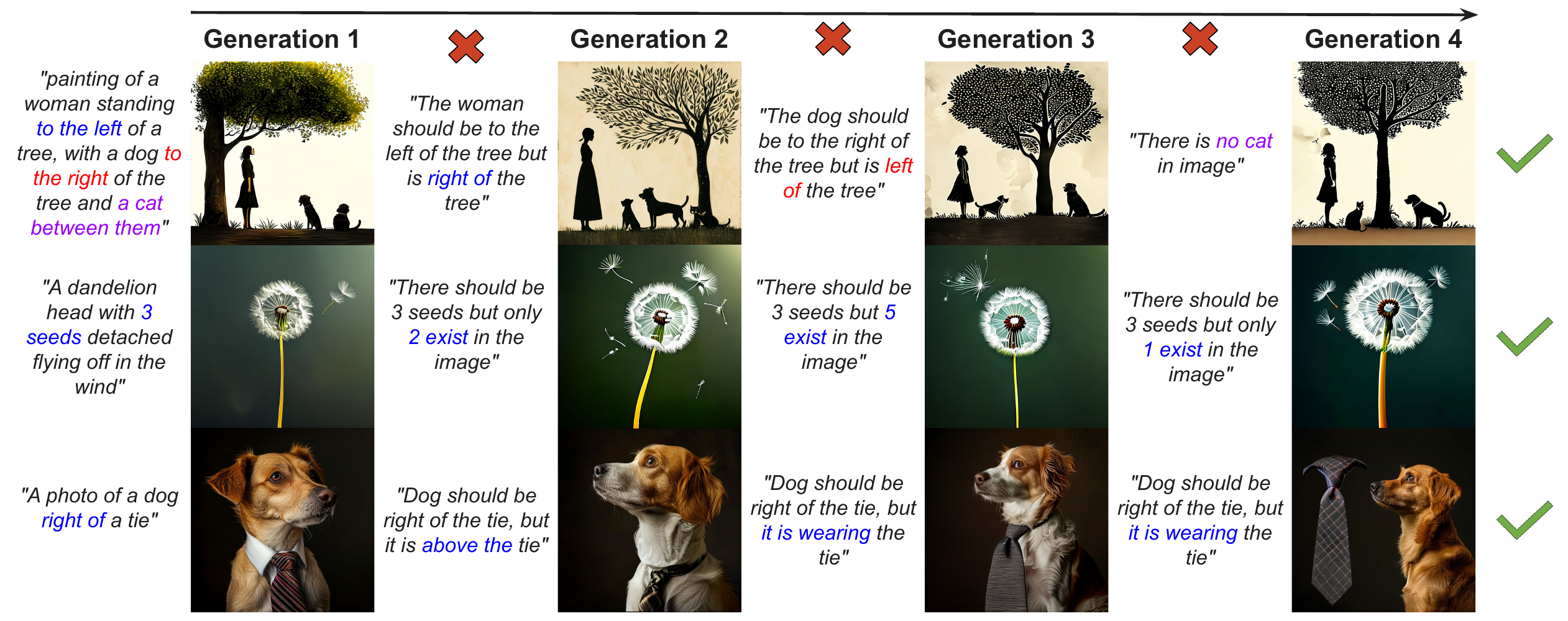}
    \caption{\textbf{Illustration of the iterative refinement process in \ours}. \ours~starts with an initial image generated from the prompt and progressively refines it based on textual feedback until the final output meets the desired criteria, demonstrating the effectiveness of our reflection-based approach. In the first sequence, \ours~handles a complex scene by gradually repositioning multiple objects—``woman," ``tree," ``cat," and ``dog"—to achieve correct spatial alignment. Additionally, it recognizes subtle object misclassifications, such as changing the second ``dog" to a ``cat" based on feedback. The second example demonstrates a counting problem, where the model iteratively adjusts the number of detached seeds until it converges to the correct count. The final example presents a particularly challenging scenario: the prompt requires the ``dog" to be positioned to the ``\emph{right of} a tie", an unusual object to appear independently. Initially, the model misinterprets the instruction, generating a dog wearing a tie. However, through multiple refinement steps, \ours~learns to separate the objects and ultimately produces the correct spatial arrangement.}
    \label{fig:figure-feedback}
\end{figure*}

\subsection{Ablation Studies}

We measure the effect of various design choices by conducting a series of ablation studies on the number of in-context feedback, Transformer layers, and image embeddings.

\subsubsection{Number of Feedback}

\label{sec:ablation-multi-round}

One of the key innovations of \ours~is the ability to utilize multi-turn feedback. We experiment with varying amounts of feedback and find that an increase in-context feedback ($K$ in Algorithm \ref{alg:inference}) yields better performance, with the 3-feedback setup achieving the best result (\cref{tab:ablation-n-feedback}).

\begin{table}[h]
    \centering
    \begin{tabular}{lccccc}
        \toprule
         & \multicolumn{5}{c}{Number of Samples }\\
         & 4 & 8 & 12 & 16 & 20 \\
        \midrule
        Baseline & 0.705 & 0.731 & 0.743 & 0.749 & 0.751 \\
        $K=1$ & 0.738 & 0.755 & 0.761 & 0.772 & 0.766 \\
        $K=2$ & 0.743 & 0.765 & 0.765 & 0.774 & 0.781 \\
        $K=3$ & \textbf{0.776} & \textbf{0.792} & \textbf{0.799} & \textbf{0.799} & \textbf{0.807} \\
        \bottomrule
    \end{tabular}
    \caption{\textbf{Ablation study on the number of in-context image-feedback pairs (K).} } 
    \label{tab:ablation-n-feedback}
\end{table}

\subsubsection{Number of Transformer Layers}

We conducted multiple experiments with varying number of Transformer layers in the Context Transformer (\cref{tab:ablation-layers}). Results show that increasing the number of Transformer layers leads to an improvement in model performance. 

\begin{table}[h]
    \centering
    \begin{tabular}{lccccc}
        \toprule
         & \multicolumn{5}{c}{Number of Samples }\\
         & 4 & 8 & 12 & 16 & 20 \\
        \midrule
        Baseline & 0.705 & 0.731 & 0.743 & 0.749 & 0.751 \\
        +1 Layer  & 0.769 & 0.790 & 0.797 & \textbf{0.801} & 0.804 \\
        +2 Layers & \textbf{0.776} & \textbf{0.792} & \textbf{0.799} & 0.799 & \textbf{0.807} \\
        \bottomrule
    \end{tabular}
    \caption{\textbf{Ablation study on number of Transformer layers.} }
    \label{tab:ablation-layers}
\end{table}

\subsubsection{Number of Image Embeddings}
We also explore different down-sampling sizes for image embeddings (default $8\times8$) and present the results in Table \ref{tab:ablation-patch-size}. Results show that using more tokens to represent past generations leads to better performance.

\begin{table}[H]
    \centering
    \begin{tabular}{lccccc}
        \toprule
         & \multicolumn{5}{c}{Number of Samples }\\
         & 4 & 8 & 12 & 16 & 20 \\
        \midrule
        Baseline & 0.705 & 0.731 & 0.743 & 0.749 & 0.751 \\
        $4\times4$  & 0.770 & 0.779 & 0.783 & 0.786 & 0.786 \\
                $6\times6$  & 0.752 & 0.781 & 0.795 & 0.795 & 0.801 \\
        $8\times8$ & \textbf{0.776} & \textbf{0.792} & \textbf{0.799} & \textbf{0.799} & \textbf{0.807} \\
        \bottomrule
    \end{tabular}
    \caption{\textbf{Ablation study on the size of image embeddings.}}
    \label{tab:ablation-patch-size}
\end{table}

\section{Discussion}

\subsection{Connection with Instruction-Guided Image Editing}
\label{sec:comparision_with_instructpix2pix}
Several works have focused on instruction-following image editing models, \eg InstructPix2Pix \cite{Brooks_2023_CVPR} and InstructDiffusion \cite{geng2024instructdiffusion}. These models take an input image and natural language instruction and perform the desired edit on the image. Compared to these methods, \ours~has two key advantages. First, our training data relaxes the strict requirement for paired input-edited images. Image-editing data consists of paired images: an input and a corresponding edited version that adheres to the instruction while maintaining visual consistency. \ours~only requires a ``good" image that avoids a problem found in a ``bad" image, which can be easily collected. Second, but more importantly, \ours~uniquely leverages multi-round feedback context and progressively refines its generations. Our results in Section \ref{sec:ablation-multi-round} demonstrate that iterative in-context feedback significantly improves performance. Figure \ref{fig:figure-feedback} also illustrates that multi-round feedback enables the progressive refinement of generations. Although some identified defects may not be fully resolved in a single iteration, the generated images converge towards a correct image after multiple rounds. The model learns this ability despite the random sampling of in-context feedback during training.


\section{Conclusion}
In this work, we presented \ours, a framework for inference-time scaling of text-to-image diffusion models that leverages reflection on past generations and natural language feedback. Results show that \ours~significantly outperforms naive best-of-N sampling, which to date has been the most effective inference-time scaling method, and establishes a new state-of-the-art on the GenEval benchmark, outperforming larger models with substantially higher sampling budgets at test-time. While the \ours~framework has shown to be a promising approach to improve text-to-image generation, it is important to note that our method inherits the biases and shortcomings of both the base text-to-image model and the VLM judge that guides refinement. For example, the VLM judge, like other VLMs, may generate incorrect feedback due to hallucination. We also find it challenging for the VLM to recognize small objects. Future work should focus on auditing these limitations and developing safeguards to ensure responsible deployment of \ours~in applications. Further discussion of limitations is provided in Appendix \ref{sec:appendix-limitation}.
\newpage
{
    \small
    \bibliographystyle{ieeenat_fullname}
    \bibliography{main}
}

\clearpage
\appendix

\section{Additional Results and Discussions}

\subsection{Generalizability of \ours}
In the main paper, we provide human evaluation results on PartiPrompts to demonstrate the generalizability of \ours~beyond GenEval-style prompts. While human evaluators are the ideal judges for text-to-image generation, we recognize that human evaluations can be expensive. In this section, we report additional results on DPG-Bench \cite{hu2024equip} to highlight the effectiveness of \ours~on a broad range of prompts.  

Table \ref{tab:appendix-dpg-bench} presents the performance of \ours~on DPG-Bench \cite{hu2024ella}. We note that several frontier open-sourced models achieve similar performance on this benchmark, in the range of 83.0-85.0. The performance gap among models on DPG-Bench is less pronounced compared to GenEval. Analysis of SANA-1.0 outputs indicates that approximately 75\% of the prompts are less challenging, as indicated by the base model achieving a score above 0.8 without inference-time scaling. This may explain why SANA-1.5 \cite{xie2025sana2} reported inference-time scaling results only on GenEval. To further illustrate the effectiveness of \ours, we construct two challenging datasets by subsampling prompts where the SANA-1.0-1.6B base model scores poorly in the single-sample setup. Specifically, we create a subset of 246 prompts (Hard-246) consisting of prompts on which the base model obtained a score below 0.8 in the single-sample setup, and a subset of 56 prompts (Hard-56) consisting of prompts on which the base model obtained a score below 0.5 in the single-sample setup. We compare with the base model and best-of-N sampling on the two subsets as well as the full benchmark. \ours~achieves better performance on all three datasets, with more pronounced differences on the two hard subsets. The results on DPG-Bench, together with the main results on GenEval and human evaluations on PartiPrompts, demonstrate the effectiveness of \ours~across diverse text-to-image generation tasks.

\begin{table}[h]
    \centering
    \begin{tabular}{c|c |ccc}
         &\textbf{ Parm.} & \textbf{All} & \textbf{Hard-246 }& \textbf{Hard-56 } \\
         \hline
         SD3-Medium & 2B & 84.1& - & -  \\
         DALLE 3 & - & 83.5 & - & - \\
            FLUX-dev & 12B & 84.0 & - & - \\
         SANA-1.5 $\dag$ & 4.8B & 85.0 & - & - \\
         \hline
         SANA-1.0 & 1.6B & 84.1& 56.1 & 24.2 \\
          +Best-of-20 & 1.6B & 85.6 &  63.4 & 41.0 \\
        +\ours & 1.7B & \textbf{86.1} & \textbf{69.5} & \textbf{51.8} \\
         \textbf{ ($\Delta$ vs Base.)} & - & \textbf{\textgreen{+2.0}} & \textbf{\textgreen{+13.4}} & \textbf{\textgreen{+27.6}} \\
         \hline
    \end{tabular}
    \caption{\textbf{Additional quantitative results on DPG-Bench \cite{hu2024ella}} $\dag$ SANA-1.5 only reported inference-time scaling (best-of-2048) on GenEval benchmark and has not been open sourced. We cite their single-sample result here.}
    \label{tab:appendix-dpg-bench}
\end{table}
\subsection{Inference Speed}
\label{sec:appendix-speed}
We benchmarked the inference speed of \ours~against the best-of-N baseline and found no significant difference in performance. Overall, \ours~ and best-of-N sampling achieved similar throughput: 11.32 samples per minute for \ours~and 10.12 samples per minute for best-of-N, where each sample includes a generated image and corresponding text feedback. Conceptually, generating N samples using \ours~has a similar computational cost to the best-of-N baseline, as both involve generating N images and running the VLM model N times. The only extra overhead comes from (1) encoding the images and text in the context and (2) computing cross-attention with extra keys and values. Furthermore, step (1) can be amortized across the denoising steps, as the context needs to be encoded only once per generated image at the beginning of the denoising loop. For N=20, the end-to-end latency is 118.57 seconds for \ours~and 105.98 seconds for the best-of-N baseline. Of the total time, 75.5\% is spent generating images with the DiT and 24.5\% is used for VLM inference.


\subsection{Connection with Self-Correcting T2I Agent}
\label{sec:appendix-agent}
Prior to the recent interest in inference-time scaling, several works attempted to achieve self-verification and correction through an agentic framework, such as SLD \cite{wu2024self} and GenArtist \cite{wang2024genartist}. These works employ a frontier LLM or VLM (e.g. GPT4) to control a set of external tools such as object detectors, segmentation models and inpainting models to verify the accuracy of text-to-image (T2I) generation and apply corrective operations to the generated image. These approaches suffer from key scalability and flexibility issues.

In terms of scalability, calling proprietary APIs for each inference is expensive. Additionally, generating function calls auto-regressively and executing multiple models per refinement round introduce significant computational overhead and latency. In contrast, \ours~and recent works on inference-time scaling only require a, significantly smaller, VLM judge model to simply generate concise per-image feedback in natural language.

In terms of flexibility, the success of these agentic frameworks depends on all submodules executing successfully, giving rise to two main problems. First, these submodules may not be up-to-date. For example, the inpainting and image editing models they use are primarily based on SDv1.5 \cite{rombach2022high} or SDXL \cite{podell2023sdxl}, resulting in suboptimal generation quality. Updating all tools to the latest architectures and base models is non-trivial, since the developers of these tools may discontinue maintenance, which is not unrealistic for most research projects. Additionally, adapting a system with numerous components to custom use cases can be challenging. For example, if a user wants to generate a painting, the pretrained object detector and segmentation models may fail on out-of-distribution cases such as painting generation. Collecting a detection and segmentation dataset and fine-tuning the object detector and segmentation model can be expensive and challenging, not to mention the difficulty of data collection for others tools like inpainting and image-editing models. In contrast, \ours~and recent inference-time scaling methods can easily adapt to new use cases as long as a judge model provides feedback, which can be obtained by fine-tuning a strong foundational VLM on limited data. In fact, we show in Figure \ref{fig:figure-feedback} (main paper) that \ours~can adapt to novel use cases such as painting generation in a zero-shot manner due to the inherent generalizability of VLMs, highlighting the flexibility of our approach.

We find these works interesting but tangential. In our early explorations, we attempted to reproduce the findings of GenArtist \cite{wang2024genartist} (NeurIPS 2024) and test its performance on the GenEval benchmark. Unfortunately, the model fails to complete the official demo script using the default prompt, as the latest version of GPT-4 generates ill-formed function calls approximately 20 seconds into the agentic loop. Our experience further highlights the inflexibility and inconsistency of these methods.

\subsection{Connection with Reinforcement Learning (RL) }

If we consider the consecutive, non-i.i.d. generative process of multiple image samples as a policy optimization problem, then \ours's training objective can be viewed as equivalent to imitation learning, where we directly apply the SFT objective to a set of target ``good actions",\ie accurate image generations. We also explored reinforcement learning objectives such as D3PO \cite{yang2023using} and Diffusion-DPO \cite{wallace2024diffusion}, which incorporate negative samples during training. However, we encountered issues with training stability. We achieved state-of-the-art results using only the SFT objective and leave further exploration of RL objectives to future work. Our results mirror those of DeepSeek-R1 \cite{guo2025deepseek}, where the authors showed that smaller LLMs can achieve substantial performance gains solely through SFT on high-quality reasoning trajectories generated by a larger model, without requiring reinforcement learning.

\subsection{Concurrent Works}

Following the success of test-time scaling in the language domain, the community has shown growing interest in applying it to text-to-image generation. Concurrently with this work, SANA-1.5 \cite{xie2025sana2} explored best-of-N sampling on a state-of-the-art DiT. Our proposed \ours~outperforms SANA-1.5, which uses 2048 samples (best-of-2048), with only 20 samples by leveraging a reflection mechanism. Also concurrent, FK-steering \cite{singhal2025general} proposed a novel latent-space search method that extends beyond random search. However, its implementation is limited to the DDIM sampler and is not easily adaptable to multi-step solvers such as the DPM-Solver++ which is used by SANA. In contrast, \ours~has a constant memory footprint, making it more scalable. After testing the official SDXL-based implementation, we find that FK-Steering causes out-of-memory errors at 20 particles. While their results are promising, we believe it has considerable room for improvement, particularly in adapting to state-of-the-art DiTs and optimizing memory usage. Another concurrent work \cite{guo2025generateimagescotlets} explored generating images using chain-of-thought (CoT) \cite{wei2022chain} reasoning and incorporates elements of self-verification and iterative improvement. However, their work focuses on autoregressive image generation models and is a direct adaptation of analogous approaches in the language domain.
 
\section{Limitations}
\label{sec:appendix-limitation}
\begin{figure}
    \centering
    \includegraphics[width=1.0\linewidth]{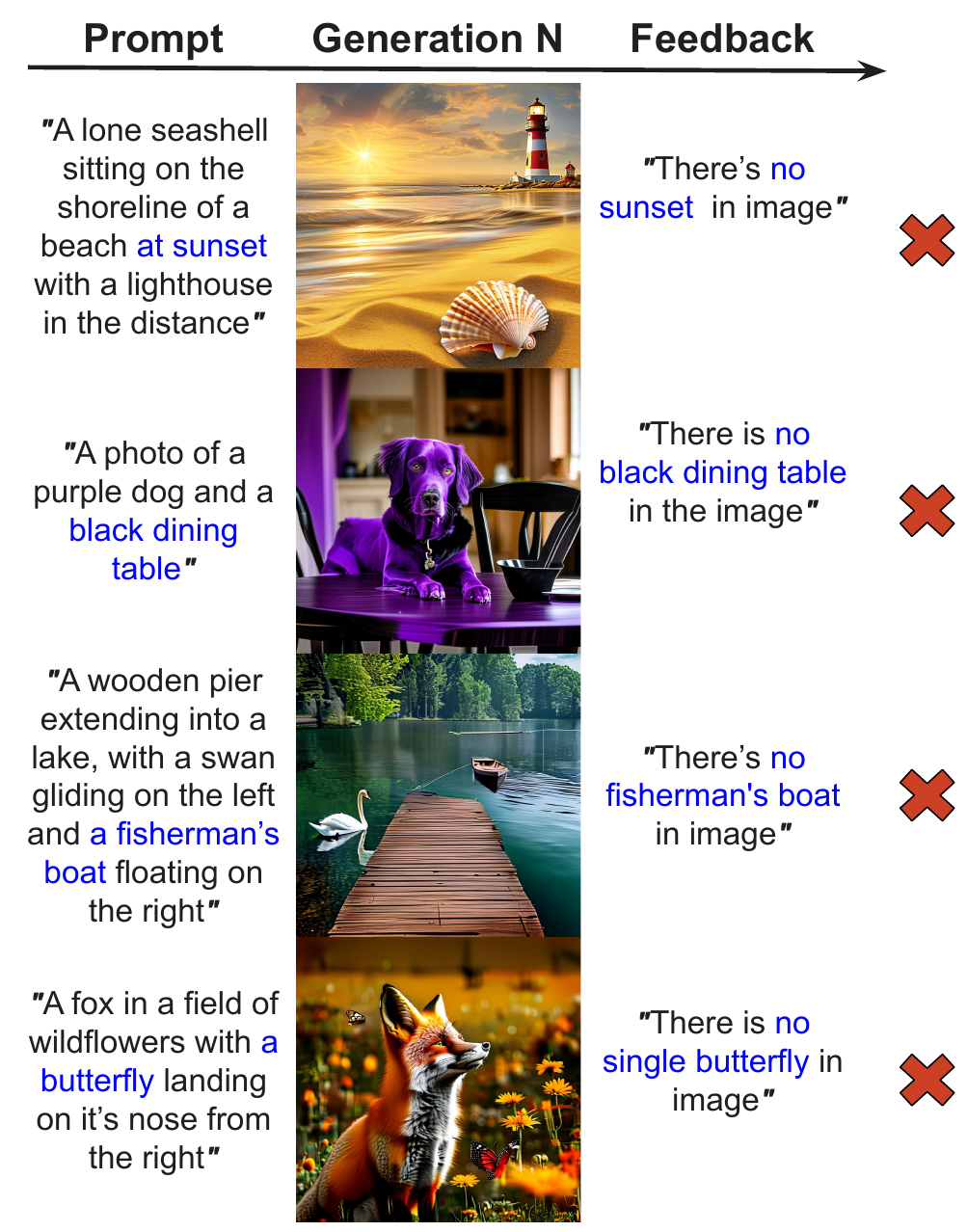}
    \caption{\textbf{Failure cases of \ours}. Failure cases of \ours. While \ours~demonstrates strong refinement capabilities, the generated feedback can occasionally introduce errors between iterations. In the first example, the model fails to recognize that the specific lighting conditions signify a ``sunset", leading to an incorrect adjustment. Similarly, in the second example, the model struggles to distinguish the color of the ``dining table" because the purple hue from the ``dog" reflects off the table, creating ambiguity. These cases highlight subjectivity in the VLM evaluation, where the model's interpretation may still be reasonable. However, the final two examples illustrate more typical failure cases. In both images, objects (``boat" and ``butterfly") are completely overlooked by the feedback model. This issue likely arises because the objects are too small or unusually shaped, which makes them difficult to detect, resulting in incorrect evaluations.}
    \label{fig:appendix-figure-failure}
\end{figure}

While we have demonstrated \ours’s effectiveness across various applications, we acknowledge its limitations. First, the training data used for the VLM judge primarily focuses on prompt alignment, e.g. whether there are sufficient objects, whether they satisfy positional constraints, \etc. Thus, the VLM judge may not be able to provide proper feedback or suggest improvements for other aspects, such as the aesthetics of an image. In general, natural language feedback datasets of this kind are more difficult to collect. We hope future datasets can help address this issue. Second, we observe that the VLM judge suffers from hallucinations, similar to its base model. We present examples of these errors in Figure \ref{fig:appendix-figure-failure}. For example, in row 3 of Figure \ref{fig:appendix-figure-failure}, the VLM mistakenly claims that the boat is not present in the image, despite the boat being clearly present and the image being correct. Lastly, we observed that the diffusion model may fail to address certain forms of feedback in a single iteration. In some cases, we observe the model iteratively refining the generation toward correctness, though it takes multiple iterations for the image to become fully aligned with the prompt. For example, in Figure \ref{fig:figure-feedback} (main paper), row 3, we observe that the position of the dog and the tie gradually move toward the desired layout. In other cases, the progression is less interpretable. For example, in Figure \ref{fig:figure-feedback} (main paper), row 2, the generated image should contain three seeds, but it undergoes an inconsistent progression of 2-5-1-3. Empirically, we observe that \ours~can generate accurate, text-aligned images with fewer inference-time samples, achieving a 22\% improvement on the counting subcategory of GenEval (Table \ref{tab:results-geneval} in main paper).
 
\section{Technical Details}

\subsection{VLM Training}
\label{sec:appendix-vlm-training}
\begin{table}[]
    \centering
    \begin{tabular}{p{0.9\linewidth}H}
        \hline
        \textbf{VLM Training Data Template} &  \\ \hline
        \texttt{\{ } \newline
        \texttt{"from": "human",} \newline
        \texttt{"value": "<image>\textbackslash{}n} \newline
        \texttt{Please evaluate this generated image based on the following prompt: [[prompt]].} \newline
        \texttt{Focus on text alignment and compositionality."} \newline
        \texttt{\},} \newline
        \texttt{\{} \newline
        \texttt{"from": "gpt",} \newline
        \texttt{"value": "[[feedback\_text]]"} \newline
        \texttt{\}} &  \\ \hline
    \end{tabular}
    \caption{VLM Training Data Template}
    \label{tab:appendix-vlm-template}
\end{table}

\begin{table*}[]
    \centering
    \begin{tabular}{c|cccc}
        \textbf{Hyperparameters} & \textbf{VLM Judge} & \textbf{\ours} & \textbf{SFT} & \textbf{Diffusion-DPO} 
        \\
         \hline
        Learning Rate &  1e-5 & 1e-5 & 1e-5 & 1e-5 \\ 
  
        Batch Size & 48 & 48 & 48 & (24, 24)* \\

        Weight Decay & 0.1 & 0 & 0 & 0 \\ 
        Optimizer & AdamW & CAME & CAME & CAME \\ 
        Schedule & 1 epoch & 5k step &  5k step & 5k step \\
        Warmup steps & 0.03 epoch & 500 step & 500 step & 500 step \\
    \end{tabular}
    \caption{\textbf{Hyperparameters used for each experiment.} * We use 24 positive samples and 24 negative samples per batch. }
    \label{tab:appendix-hyperparameters}
\end{table*}

Following SANA-1.5 \cite{xie2025sana2}, we format the VLM training data into a conversation format. Our template differs from SANA because we use a different base model, Qwen-2.5-VL 3B \cite{bai2025qwen2}. We present the template in Table \ref{tab:appendix-vlm-template}. We provide hyperparameters of our training run in Table \ref{tab:appendix-hyperparameters}. 

\subsection{Diffusion Transformer}

\subsubsection{Vision-Encoder}

The vision encoder is a SigLIP-Large \cite{zhai2023sigmoid} that encodes each image into a feature map of size $24 \times 24 = 1024$.The feature map is then downsampled to $8 \times 8 = 64$ via average pooling and flattened into a 1D sequence of length $64$. We then use a two layer MLP with GELU activation to project the features to match the input dimension of the Context Transformer. To improve training stability, we add an RMSNorm layer after the projector. Before training, we freeze the SigLIP model. The projector is trained end-to-end with the rest of the DiT.

\subsubsection{Text-Encoder}

We use Gemma-2B \cite{team2024gemma} as the text encoder for text feedback. It is kept frozen during training. Since Gemma-2B is also used by SANA as the prompt encoder, no additional parameters are introduced to the overall system.

\subsubsection{Context Transformer}

The Context Transformer is a two-layer Transformer. Its primary purpose is to (1) align encoded features with the features space of the base DiT and (2) associate the feedback with the corresponding image. Each Context Transformer consists of a standard Transformer block, including a self-attention layer followed by a feed-forward network. We use the exact FFN design of Qwen2.5-VL \cite{bai2025qwen2}. For the self-attention layer, we incorporated rotary positional embeddings \cite{heo2024rotary} following the design of many modern LLMs and VLMs. 

\subsubsection{Training}

We report the training hyperparameters for \ours, SFT, and Diffusion-DPO baselines in Table \ref{tab:appendix-hyperparameters}. We use the CAME optimizer \cite{luo-etal-2023-came} to train the DiT, following the approach in SANA \cite{xie2025sana2}. For Diffusion-DPO, we tested three values of $\beta$, the hyperparameter controlling the KL divergence penalty, and determined that $\beta=2000$ produces the optimal result.

\subsection{Human Evaluation Details}
\label{sec:appendix-human-eval}
\begin{figure}
    \centering
     \fbox{\includegraphics[width=1\linewidth]{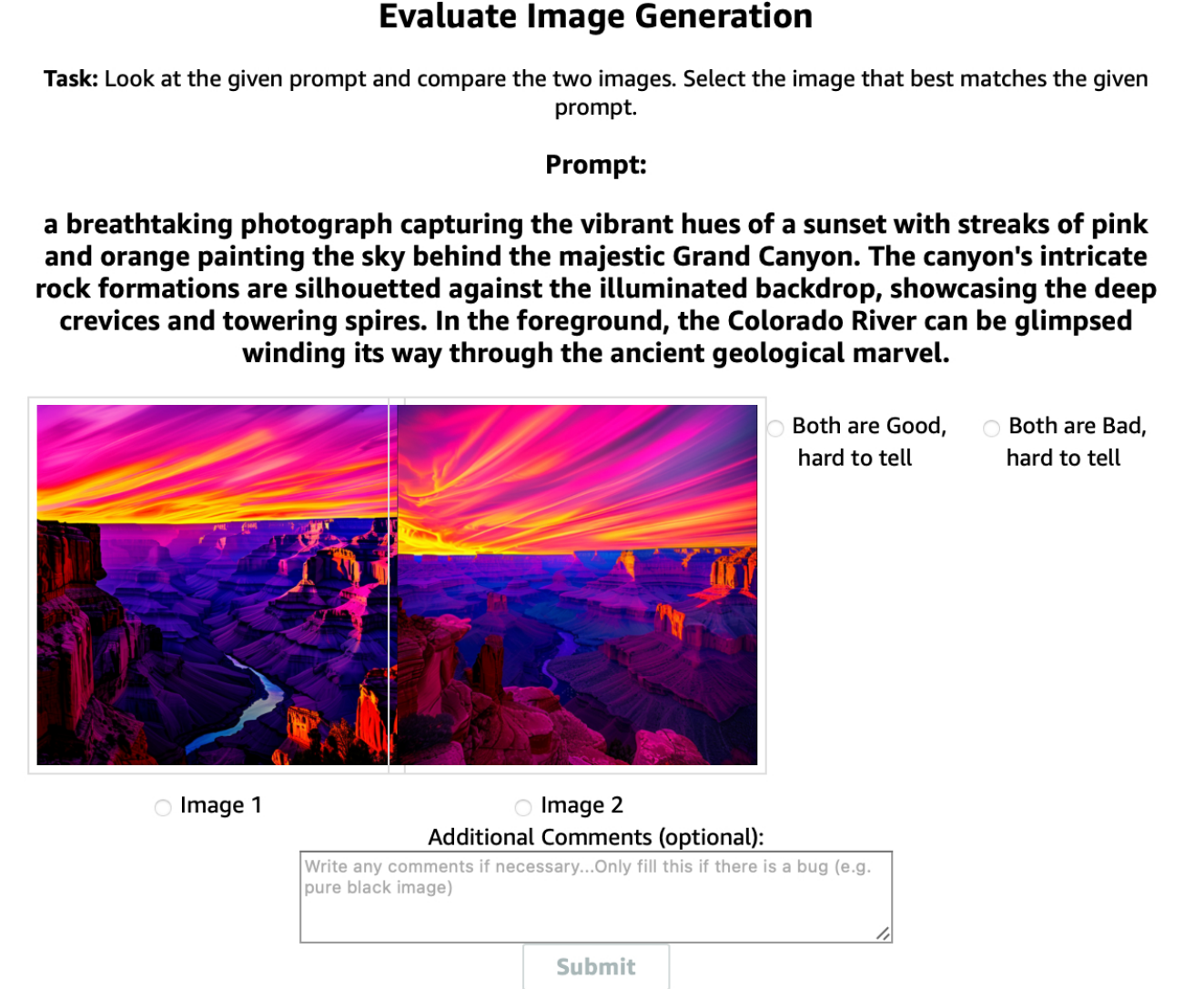}}
    \caption{\textbf{User interface for human annotators.}}
    \label{fig:appendix-human-interface}
\end{figure}

We use Amazon Mechanical Turk for human evaluations. We present the user interface provided to human annotators in Figure \ref{fig:appendix-human-interface}. We collect three evaluations per image pair and compared \ours(N=20) with best-of-20 for each prompt. We randomly selected 100 prompts from the PartiPrompts dataset and generated 100 corresponding image pairs for human annotators. In total, 300 annotations were collected.

\section{Additional Qualitative Examples}
\begin{figure*}[t]
    \centering
    \includegraphics[width=1.0\linewidth]{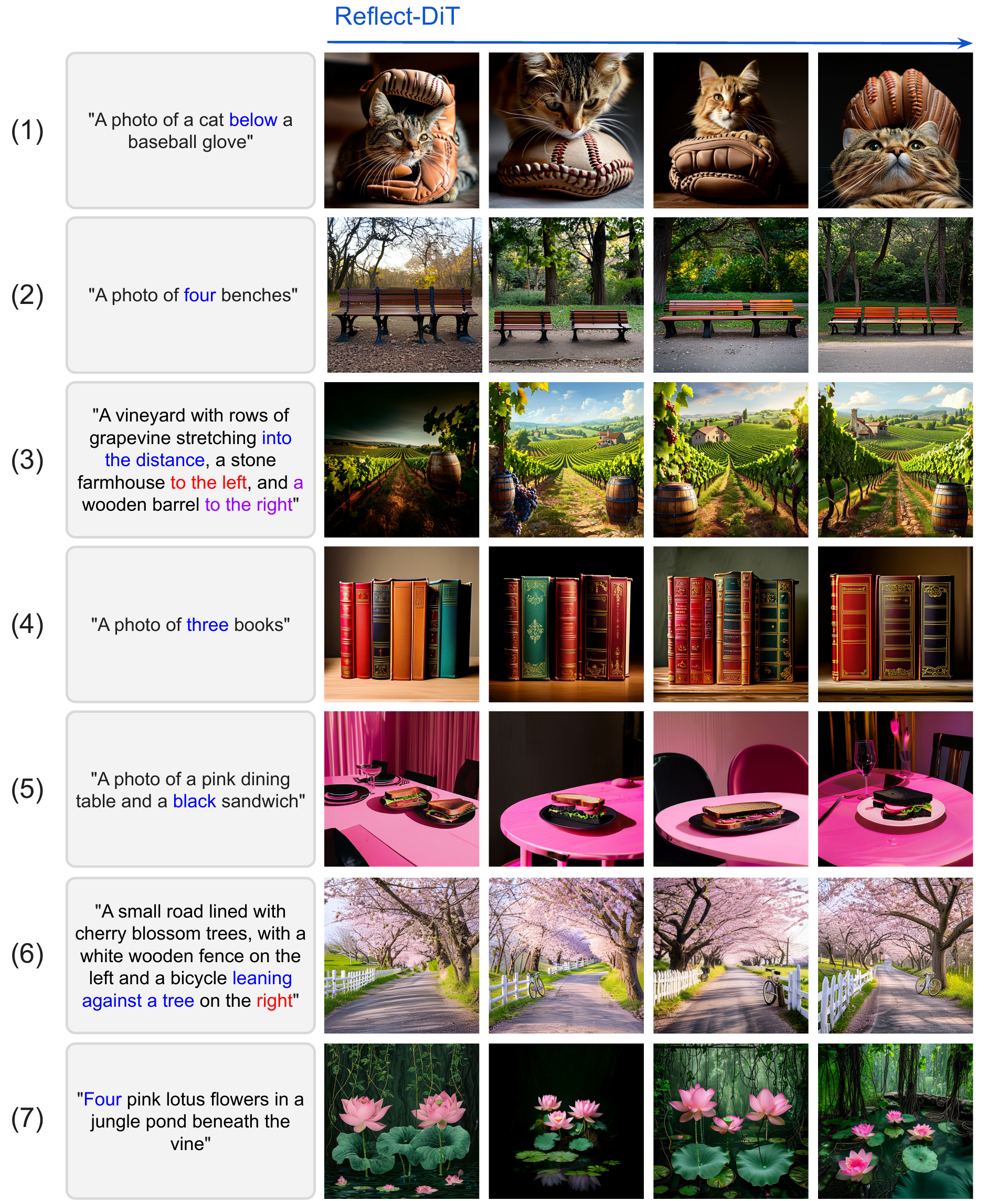}
    \caption{\textbf{Additional qualitative examples from \ours.}}
    \label{fig:appendix-figure-extra}
\end{figure*}

We present additional evaluation results in Figure \ref{fig:appendix-figure-extra}. Examples 1 and 6 demonstrate how \ours~guides the generation process to accurately position objects within a scene. Examples 2, 4, and 7 focus on object counting, ensuring that the correct number of distinct items. Example 3 presents a particularly complex prompt, where \ours~accurately positions all objects while maintaining the correct quantity, such as the specified number of ``wooden barrels". Lastly, Example 5 highlights a challenging case—separating object identity from color attributes—that many generative models struggle with. Typically, models often conflate color and object identity, making requests like "a black sandwich" difficult to fulfill. However, \ours~successfully distinguishes these attributes, demonstrating its advanced capability to handle nuanced prompts.

\section{Reproducibility Statement}
We will release the training code and data for the DiT and VLM judge model, and pretrained checkpoints. We will also release the generated images that produce the main result on GenEval benchmark. Additionally, we will release the list of prompts in Hard-246 and Hard-56 subset of DPG-Bench.

\end{document}